\documentclass[sigconf,9pt,screen=false,prologue,table,usenames,dvipsnames,svgnames,x11names, nonacm]{acmart}

\usepackage{epstopdf}
\usepackage{booktabs}
\usepackage[utf8]{inputenc}
\usepackage[T1]{fontenc}
\usepackage{xcolor}
\usepackage{pifont}
\usepackage{subcaption}
\usepackage{balance}
\usepackage[linesnumbered,ruled,vlined, noend]{algorithm2e}
\let\oldnl\nl
\newcommand{\nonl}{\renewcommand{\nl}{\let\nl\oldnl}}

\SetCommentSty{mycommfont}
\usepackage{multirow}
\usepackage[abbreviations]{foreign}

\usepackage{pifont}
\usepackage{comment}
\newcolumntype{P}[1]{>{\centering\arraybackslash}p{#1}}
\newcolumntype{M}[1]{>{\centering\arraybackslash}m{#1}}
\usepackage{listings}
\usepackage{svg}
\usepackage{amsmath}
\usepackage{tabularx, makecell, booktabs}
\newcommand\mydots{\makebox[0.8em][c]{.\hfil.\hfil.}}
\AtBeginDocument{

  \newcommand{\sys}{\textsc{Pelta}\xspace}
  \definecolor{darkgreen}{rgb}{0.3,0.5,0.3}
  
  \definecolor{darkred}{rgb}{0.5,0.3,0.3}

}

\usepackage{framed}
\colorlet{shadecolor}{Azure2}

\usepackage{cleveref} 
\crefname{section}{§}{§§}

\usepackage[binary-units,per-mode=symbol]{siunitx}

\usepackage[inline]{enumitem}
\setlist{noitemsep,topsep=0pt,parsep=0pt,partopsep=0pt}
\usepackage{setspace}
\usepackage[margin=5pt,font={stretch=0.9}]{caption}

\usepackage{hyperref}
\hypersetup{
 hidelinks = true,
    colorlinks = false,
    linkbordercolor = {white}
}

\ccsdesc[500]{Computing methodologies~Machine learning}
\ccsdesc[500]{Security and privacy}

\ccsdesc[500]{Computing methodologies~Machine learning}

\setcopyright{none}
\renewcommand\footnotetextcopyrightpermission[1]{} 
\makeatletter
\renewcommand\@formatdoi[1]{\ignorespaces}
\makeatother

\begin{document}

\title[Pelta: Shielding Transformers to Mitigate Evasion Attacks in Federated Learning]{Pelta: Shielding Transformers to Mitigate\\Evasion Attacks in Federated Learning}

\author{Simon Queyrut}
\orcid{0000-0002-1354-9604}
\affiliation{
  \institution{University of Neuchâtel}
  \city{Neuchâtel}
  \country{Switzerland}
}
\email{simon.queyrut@unine.ch} 

\author{Yérom-David Bromberg}
\orcid{0000-0002-3812-3546}
\affiliation{
  \institution{University of Rennes 1 - Inria - IRISA}
  \city{Rennes}
  \country{France}
}
\email{david.bromberg@irisa.fr}

\author{Valerio Schiavoni}
\orcid{0000-0003-1493-6603}
\affiliation{
  \institution{University of Neuchâtel}
  \city{Neuchâtel}  
  \country{Switzerland}
}
\email{valerio.schiavoni@unine.ch}

\definecolor{darkgreen}{rgb}{0.3,0.5,0.3}
\definecolor{darkblue}{rgb}{0.3,0.3,0.5}
\definecolor{darkred}{rgb}{0.5,0.3,0.3}

\newboolean{showcomments}
\setboolean{showcomments}{false}
\ifthenelse{\boolean{showcomments}}{ 
  \newcommand{\mynote}[3]{
    \fbox{\bfseries\sffamily\scriptsize#1}
    {\small$\blacktriangleright$
     \textsf{\emph{\color{#3}{#2}}}$\blacktriangleleft$
     }
   }
}{\newcommand{\mynote}[3]{}}
\newcommand{\sq}[1]{\mynote{simon}{#1}{orange}}
\newcommand{\vs}[1]{\mynote{valerio}{#1}{red}}

\newcommand{\old}[1]{\mynote{Old}{#1}{red}}
\newcommand{\new}[1]{\mynote{New}{#1}{blue}}

\begin{abstract}
The main premise of federated learning is that machine learning model updates are computed locally, in particular to preserve user data privacy, as those never leave the perimeter of their device.
This mechanism supposes the general model, once aggregated, to be broadcast to collaborating and non malicious nodes.
However, without proper defenses, compromised clients can easily probe the model inside their local memory in search of adversarial examples.
For instance, considering image-based applications, adversarial examples consist of imperceptibly perturbed images (to the human eye) misclassified by the local model, which  can be later presented to a victim node's counterpart model to replicate the attack.
To mitigate such malicious probing, we introduce \sys, a novel shielding mechanism leveraging trusted hardware. 
By harnessing the capabilities of Trusted Execution Environments (TEEs), \sys masks part of the back-propagation chain rule, otherwise typically exploited by attackers for the design of malicious samples.
We evaluate \sys on a state of the art ensemble model and demonstrate its effectiveness against the Self Attention Gradient adversarial Attack. 

\end{abstract}

\keywords{federated learning, evasion attacks, mitigations, trusted execution environment}


\maketitle
\section{Introduction}
With the proliferation of edge devices and small-scale local servers comes an ever increasing trove of data. 
Examples include smart homes, medical infrastructures, mobile phones, \etc. 
A common trend is to deploy and execute data-driven machine learning (ML) algorithms directly at edge on such devices for a large diversity of purposes.
However, without proper mechanisms, users are by default set to share their data for the training process.
As an alternative and to protect user privacy when training a ML model, federated learning (FL)~\cite{45648,kairouz2021advances} lets non-malicious clients send to a trusted server their individual updates of a shared model.
In turn, the trusted server aggregates such updates, before broadcasting the updated model to each clients.
This approach prevents the user-data from ever leaving the user node from a given device.
Also, the updated model is used to compute the next update with a fresh batch of local data, as soon as it is received by each client.

\begin{figure}[!t]
    	\begin{center}
		\includegraphics[scale=0.6,trim={0 4 0 0}]{{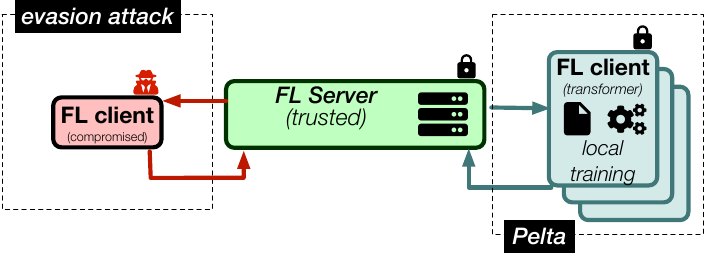}}
    	\end{center}
		\vspace{-5pt}
	\caption{\label{fig:fl}Federated learning under trusted and compromised nodes. \sys shields against evasion attacks.}
	\vspace{-14pt}
\end{figure}


Transformers~\cite{NIPS2017_3f5ee243} are popular multi-purpose deep learning (DL) models.
They harness the so-called \textit{self-attention mechanism}, which weighs the relative positions of tokens inside a single input sequence to compute a representation of that given sequence. 
Transformers create a rich and high-performing continuous representation of the whole input text or image~\cite{Raganato2018AnAO}, effectively used across a diverse range of tasks, yielding state of the art results in several areas, \eg language modeling~\cite{Shoeybi2019, brown2020language}, translation~\cite{DBLP:journals/corr/abs-2104-06022}, audio classification~\cite{NEURIPS2021_76ba9f56}, computer vision tasks such as image classification with Vision Transformer models (ViT)~\cite{50650}, object detection \cite{wei2022FD}, semantic segmentation \cite{beit}, \etc.
In the FL training context, both the fine tuning of Transformer-based models pre-trained on large-scale data and the training of their lightweight counterparts have sparked interest in industry and academia~\cite{ro2022scaling}.

In this work, we assume an FL scenario with compromised nodes (see Fig.~\ref{fig:fl}).
Because it is the hardest to defend against, we consider the \emph{white-box} setting, \ie the model's characteristics are completely known, and an attacker can leverage gradient computation inside the model to craft adversarial examples.
For instance, in the case of vision classification, an attacker leverages the model's gradients to craft images designed to fool a classifier while appearing seemingly unaltered to humans, launching a so-called gradient-based \emph{evasion} (or \emph{adversarial}) \emph{attack}~\cite{DBLP:journals/corr/GoodfellowSS14, 10.1007/978-3-642-40994-3_25, DBLP:journals/corr/SzegedyZSBEGF13}.\footnote{While we use Vision Transformers for illustration and description purposes, the mitigation mechanisms presented and implemented later are directly applicable to other classes of models including DNNs or Transformers, such as those for NLP, audio processing, \etc.}
In a FL context. compromised clients can probe their own device memory to launch an evasion attack against ViT. 
In a white-box scenario, these attacks exploit the model in clear at inference time, by perturbing the input and having it misclassified. Such perturbations are typically an additive mask crafted by following the collected gradients of the loss w.r.t the input pixels and applying it to the input signal. 

In this paper, we propose \sys: it is a defense that mitigates gradient-based evasion attacks on a client node launched during inference by leveraging hardware obfuscation to hide a few in-memory values, \ie those close to the input and produced by the model during each pass. 
This leaves the attackers unable to complete the chain-rule of the back-propagation algorithm and compute the perturbation to update his adversarial sample.
To this end, we rely on hardare-enabled trusted execution environments (TEE), by means of enclaves, secure areas of a processor offering privacy and integrity guarantees.
Notable examples include Intel SGX~\cite{costan2016intel} or Arm TrustZone~\cite{amacher2019performance}. 
TEEs can lead to a restriced white-box scenario, \ie, a stricter setting in which the attacker cannot access absolutely everything from the model he seeks to defeat,
hence impairing his white-box attack protocols designed for the looser hypothesis.
In our context, we deal specifically with TrustZone, given its vast adoption, performance~\cite{amacher2019performance} and support for attestation~\cite{menetrey2022watz}.
However, TrustZone enclaves have limited memory (up to 30~MB in some scenarios), making it challenging to completely shield the state of the art Transformer architectures often larger than 500~MB. This constraint therefore calls for such a hardware defense to be a light, partial obfuscation of the model.

Our main contributions are as follows:
\begin{enumerate}[label=(\roman*)]
    \item We show that it is possible to mitigate evasion attacks during inference in a FL context through hardware shielding.
	\item We describe the operating principles of \sys, our lightweight gradient masking defense scheme.
    \item We apply \sys to shield layers of a state of the art ensemble model against the Self-Attention blended Gradient Attack to show the scheme effectively provides high protection in a nigh white-box setting.
\end{enumerate}

The rest of the paper is as follows. 
\S\ref{section:related} surveys related work.
\S\ref{section:threat} presents our threat model. 
\S\ref{section:design} describes the principles of the \sys shielding defense. 
We evaluate it on an ensemble model against a gradient-based attack and discuss the results in \S\ref{section:eval}. 
Finally, we conclude and discuss future work in \S\ref{section:conclu}.
\section{Related Work}
\label{section:related}
A significant body of work exists towards defending against adversarial attacks in a white-box context~\cite{CarliniSurvey}. 
Because of its few-constraints hypothesis, particular endeavour is directed towards refining adversarial training (AT) methods~\cite{DBLP:journals/pr/QianHWZ22, DBLP:journals/corr/abs-2102-01356}.
However, recent studies show a trade-off between a model's generalization capabilities (\ie, its standard test accuracy) and its robust accuracy~\cite{raghunathan*2019adversarial, Nakkiran2019AdversarialRM, tsipras2018robustness, Chen2020MoreDC}. 
AT can also expose the model to new threats~\cite{https://doi.org/10.48550/arxiv.2202.10546} and, perhaps even more remarkably, increase robust error at times~\cite{DBLP:journals/corr/abs-2203-02006}. 
It is possible to use AT at training time as a defense in the federated learning context~\cite{10.48550/ARXIV.2203.04696} but it creates its own sensitive exposure to a potentially malicious server. 
The latter can restore a feature vector that is tightly approximated when the probability assigned to the ground-truth class is low (which is the case for an adversarial example)~\cite{DBLP:conf/iclr/ZhangCSBDH19}. 
Thus, the server may reconstruct adversarial samples of its client nodes, successfully allowing for privacy breach through an \emph{inversion attack}, \ie, reconstructing elements of private data in other nodes' devices.
Surprisingly, \cite{10.48550/ARXIV.2203.04696} also uses randomisation at inference time~\cite{10.5555/3454287.3454433} to defend against iterative gradient-based adversarial attacks, even though much earlier work expresses worrying reserves about such practice \cite{DBLP:conf/icml/AthalyeC018}. 

In FL, where privacy of the users is paramount, defending against inversion is not on par with the security of the model in itself and attempts at bridging the two are ongoing~\cite{liu2022threats}. 
These attempts focus on defending the model at training time or against \emph{poisoning}, \ie, altering the model's parameters to have it underperfom in its primary task or overperform in a secondary task unbeknownst to the server or the nodes. 
On the other hand, defenses against inversion attacks have seen a surge since the introduction of FL. 
DarkneTZ~\cite{10.1145/3386901.3388946}, PPFL~\cite{10.1145/3458864.3466628} and GradSec~\cite{Messaoud2022Aug} mitigate these attacks with the use of a TEE.
By protecting sensitive parameters, activations and gradients inside the enclave memory, the risk of gradient leakage is alleviated and the white-box setting is effectively limited, thus weakening the threat.
However, \sys is conceptually different from these methods, as it does not consider the gradient of the loss \emph{w.r.t.} the parameters, but \emph{w.r.t.} the input image instead. 
The former can be revealing private training data samples in the course of an inversion attack, while the latter only ever directs the perturbation applied to an input when conducting an adversarial attack. 
Other enclave-based defenses do not focus on adversarial attacks, are based on SGX (meaning looser enclave constraints) and do not deal with ML computational graphs in general~\cite{10.1145/3488659.3493779, Hanzlik2021Jun}, yet present defense architectures of layers that fit our case~\cite{DBLP:journals/corr/abs-1807-00969}.
Where those were tested only for CNN-based architectures or simple DNNs, \sys can protect a larger, more accurate Transformer-based architecture. 
The robustness of the two types against adversarial attacks was compared in prior studies~\cite{DBLP:journals/corr/abs-2110-02797, 9710333, DBLP:journals/corr/abs-2106-03734}.

In~\cite{Zhang2021Nov}, authors mitigate inference time evasion attacks by pairing the distributed model with a node-specific \emph{attractor}, which detects adversarial perturbations, before distributing it. 
However, \cite{Zhang2021Nov} assumes the nodes only have black-box access to their local model.
Given the current limitations on the size of encrypted memory in particular for TrustZone enclaves~\cite{amacher2019performance}, it is currently unfeasible to completely shield models such as VGG-16 or larger. 
This is why \sys aims at exploring a more reasonable use case of the hardware obfuscation features of TrustZone by shielding only a subset of the total layers, as we detail more later.
It could thus be used as an overlay to the aforementioned study.

Overall, gradient obfuscations as a defense mechanism against evasion attacks have been studied in the past~\cite{DBLP:conf/icml/AthalyeC018, tramer2018ensemble}. 
Authors discuss from a theoretical perspective the fragility of relying on masking techniques.
Instead, we show that it fares well protecting a state of the art architecture against inference time gradient-based evasion attacks even by using off-the-shelf hardware with limited resources.
We further elaborate on the strong ties to the Backward Pass Differentiable Approximation (BPDA) attack~\cite{DBLP:conf/icml/AthalyeC018} in \S\ref{section:design}.

\section{Threat Model}
\label{section:threat}
We assume an honest-but-curious client attacker, which do not tamper with the FL process and message flow.
The normal message exchanges defined by the protocol are not altered. 
Attackers have full access to the model.
A subset of the layers are \emph{shielded}, under a restricted white-box setting.
A layer $l$ is shielded (as opposed to its normal \textit{clear} state) if some variables and operations that directly lead to computing gradient values from this layer are obfuscated, \ie hidden from an attacker. 
In a typical deep neural network (DNN) $f$ such that 
$$f=\operatorname{softmax} \circ f^n \circ f^{n-1} \circ \cdots \circ f^1$$
with layer $f^i=\sigma^i(W^i\cdot x+b^i)$ ($\sigma^i$ are activation functions), this implies, from left (close to the input) to right (close to loss computation):
 the input of the layer $a^{l-1}$;
 its weight $W^{l}$ and bias $b^{l}$;
 its output $z^{l}$ and parametric transform $a^{l}$.
In general terms, a layer encompasses at least one or two transformations.
The attacker probes its own local copy of the model to search for adversarial examples, ultimately to present those to victim nodes and replicate the misclassification by their own copy of the model. 
\section{Design}
\label{section:design}
We explain first the design of \sys shielding of a model, with details on the general mechanisms and the shielding approach in \S\ref{ssec:peltashield}.
Then, we confront \sys to the case of the BPDA~\cite{DBLP:conf/icml/AthalyeC018} attack in \S\ref{subsection:bpda}.
\subsection{Back-Propagation Principles and Limits}
Recall the back-propagation mechanism; consider a computational graph node $x$, its immediate (\ie, generation $1$) children nodes $\{u^1_j\}$, and the gradients of the loss with respect to the children nodes $\{{d \mathcal{L}}/{d u^1_j}\}$.
The back-propagation algorithm uses the chain rule to calculate the gradient of the loss $\mathcal{L}$ with respect to ${x}$ as in
\begin{equation}
\label{eqn:backprop}
    \frac{d \mathcal{L}}{d {x}}=\sum_j\left(\frac{\partial f^1_j}{\partial {x}}\right)^T \frac{d \mathcal{L}}{d u^1_j}
\end{equation}
where $f^1_j$ denotes the function computing node $u^1_j$ from its parents $\alpha^1$, \ie $u^1_j=f^1_j\left(\alpha^1\right)$.
In the case of Transformer encoders, $f^i_j$ can be a convolution, a feedforward layer, an attention layer, a layer-normalization step, \etc.
Existing shielded white-box scenarios~\cite{Messaoud2022Aug} protect components of the gradient of the loss \emph{w.r.t.} the parameters $\nabla _{\theta}\mathcal{L}$, to prevent leakage otherwise enabling an attacker to conduct inversion attacks.
Instead, we seek to mask gradients that serve the calculation of the gradient of the loss \emph{w.r.t.} the \emph{input} $\nabla _{x}\mathcal{L}$ to prevent leakage enabling gradient-based adversarial attacks, as those exploit the backward pass quantities (\ie, the gradients) of the input to maximise the model's loss.

Because the model shared between nodes in the FL group still needs to back-propagate correct gradients at each node of the computational graph, masking only an intermediate layer is useless, since it does not prevent the attacker from accessing the correct in-memory gradients on the clear left-hand side of the shielded layer. 
We thus always perform the gradient obfuscation of the first trainable parameters of the model, \ie, its leftmost successive layers: it is the lightest way to alter $\nabla _{x}\mathcal{L}$ through physical masking.
Specifically, using Eq.~\ref{eqn:backprop}, where $x$ denotes the input image (\ie, the adversarial instance $x_{adv}$), an attacker could perform any gradient-based evasion attack (a special case where a layer is not differentiable is discussed in \S\ref{subsection:bpda}). 
\sys renders the chain rule incomplete, by forcibly and partially masking the left-hand side term inside the sum, $\{\partial f_{j}^{1}/\partial {x}\}$, hence relying on the lightest possible obfuscation to prevent collecting these gradients and launch the attack. 
\subsection{\sys shielding}\label{ssec:peltashield}
We consider the computational graph of an ML model:
$$\mathcal{G}=\left\langle n, l, E, u^1 \ldots u^n, f^{l+1} \ldots f^n\right\rangle$$ where $n$ and $l$ respectively represent the number of nodes, \ie, transformations, and the number of leaf nodes (inputs and parameters) \emph{s.t.} $1 \leq l<n$. 
$E$ denotes the set of edges in the graph. 
For each $(j,i)$ of $E$, we have that $j<i$ with $j\in\{1 \mydots(n-1)\}$ and $i\in\{(l+1)\mydots n\}$. The $u^i$ are the variables associated with every numbered vertex $i$ inside $\mathcal{G}$. They can be scalars or vector values of any dimension. The $f^i$ are the differentiable functions associated with each non-leaf vertex.
We also assume a TEE enclave $\mathcal{E}$ that can physically and unequivocally hide quantities stored inside (\eg side-channel to TEEs are out of scope).

To mitigate the adversarial attack, the \sys shielding scheme presented in Alg.~\ref{algo:proc} stores in the TEE enclave $\mathcal{E}$ all the children jacobian matrices of the first layer, $\{\partial f_{j}^{1}/\partial {x}\}$.
\begin{algorithm}[!t]
  \SetAlgoLined\DontPrintSemicolon
  \SetKwFunction{algo}{algo}\SetKwFunction{proc}{proc}
  \SetKwProg{myalg}{Algorithm}{}{}
  \KwData{$\mathcal{G}=\langle n, l, E, u^1 \ldots u^n, f^{l+1} \ldots f^n\rangle$; enclave $\mathcal{E}$}
  \SetKwProg{myproc}{Algorithm}{}{}
  \nonl\myproc{\DataSty{\sys}$(\mathcal{G})$}{
  $S \gets$ \DataSty{Select}$(u^{l+1} \ldots u^n)$ \label{line:1}\;
  \lFor{$u$ in $S$}
        {\DataSty{Shield}$(u, \mathcal{E})$}
   \KwRet\;}
   \nonl\myalg{\DataSty{Shield$(u^i, \mathcal{E})$}}{
   $\mathcal{E} \gets \mathcal{E}+\{u^i\}$ \label{line:6}\;
   $\alpha^i \gets \langle u^j \mid(j, i) \in E \rangle$\tcc*[f]{get parent vertices}\;
   \For{$u^j$ in $\alpha^i$} {\If{$u_j$ is input \label{line:9}}
   {$J^{j \rightarrow i} \gets {\partial f^i\left(\alpha^i\right)}/{\partial u^j}$ \tcc*[f]{local jacobian} \label{line:10}\;
        $\mathcal{E} \gets \mathcal{E}+\{J^{j \rightarrow i}\}$ \label{line:11}\;
        }
        \DataSty{Shield}$(u^j,\mathcal{E})$ \tcc*[r]{move on to parent}}
  \KwRet \label{line:13} \;}{}
  \caption{\sys shielding}
  \label{algo:proc}
\end{algorithm} 
 Note that, given the summation in Eq.~\ref{eqn:backprop}, obfuscating only some of the children jacobians $\partial f_1^1 / \partial x$, $\partial f_2^1 / \partial x \mydots$ would already constitute an alteration of $\nabla_x \mathcal{L}$. 
We however chose that \sys masks all the partial factors of this summation to not take any chances.
This obfuscation implies that intermediate gradients should be masked as well. Indeed, because they lead to $\partial f_{j}^{1}/\partial {x}$ through the chain rule, the intermediate gradients (or \textit{local jacobian}) $J^{0 \rightarrow 1}=\partial f^1\left(\alpha^1\right) / \partial u^0$ between the input $x=u^0$ of the ML model and its first transformation must be masked~(Alg.~\ref{algo:proc}-line \ref{line:11}). 
Because one layer may carry several 
transforms on its local input (\eg linear then $\operatorname{ReLU}$), local jacobians should be masked for at least as many subsequent generations of children nodes as the numbers of layers to shield.
The number of generations is directly determined at the arbitrary selection step~(Alg.~\ref{algo:proc}-line \ref{line:1}) where the defender choses how far the model should be shielded, \ie, the rightmost masked nodes. 
In practice, selecting the first couple of nodes likely induces enough alteration to mitigate the attack and prevent immediate reconstruction of the hidden parameters through either direct calculus (input-output comparison) or inference through repeated queries~\cite{Oh2019Sep}. 
From this right-hand side frontier, the defender may recursively mask the parent jacobians ~(Alg.~\ref{algo:proc}-line \ref{line:11},~\ref{line:13}).

In adversarial attacks, the malicious user keeps the model intact and treats only the input $x_{adv}$ as a trainable parameter to maximize the prediction error of the sample.
We note that the local jacobians~(Alg.~\ref{algo:proc}-line \ref{line:10}) between a child node and their non-input parents need not be hidden because the parents are not trainable (they should be input of the model to meet the condition at Alg.~\ref{algo:proc}-line \ref{line:9}).
Such quantities are, in effect, simply not present in the back-propagation graph and could not constitute a leak of the sensitive gradient.
Similarly, notice how \textsc{Select} supposes the rightmost masked nodes be chosen \emph{s.t.} they come \textit{after} every input leaf nodes (\ie, from subsequent generations). This condition: $u^i\in S\Rightarrow i>l$, insures no information leaks towards trainable input leaf nodes.
After all sensitive partials are masked by Alg.~\ref{algo:proc}, such a set of obfuscated gradients is a subset of the total forward partials that would lead to a complete chain rule and is noted $\left\{\partial f/\partial {x}\right\}^L$ assuming $L\leq n$ is the rightmost vertex that denies the attacker information to complete the chain rule.
Since all the forward partials up until $L$ are masked, then $\partial f^L /\partial {x}$ is protected and the resulting under-factored gradient vector (\ie the \textit{adjoint} of $f^{L+1}$) is a vector in the shape of the leftmost clear layer $f^{L+1}$ and noted $\delta_{L+1}=d \mathcal{L} /d {u^{L+1}}$. 
A white-box setting assumes an attacker has knowledge of both the subset $\left\{\partial f/\partial {x}\right\}^L$ of all forward partials and $\delta_{L+1}$ to perform a regular gradient-based update on his $x_{adv}$.
However in \sys, the attacker is only left with the adjoint $\delta_{L+1}$.

Finally, the forward pass quantities $u^i$ that may lead to the unambiguous recovery of the hidden set $\left\{\partial f/\partial {x}\right\}^L$ are masked (Alg.~\ref{algo:proc}-\ref{line:6}).
This insures that arguments $\alpha^i$ of the transformations $f^i$ cannot be further exploited by the attacker. 
This could happen when one node is a linear transform of the other as in $u^{i+1}=f^{i+1}((W,u^{i}))=W\times u^{i}$.
In this case, the local jacobian $J^{i \rightarrow i+1}$ is known to be exactly equal to $W$.
Notice that weights and biases of a DNN would be effectively masked, as they are regarded as leaf vertices of the model's computational graph for the $f^i((u^{i}, u^{i-1}, u^{i-2}))=u^{i-1}\cdot u^{i} + u^{i-2}=Wx + b$ operation. Similarly, the outputs of the transformations, $u^i=f^i(\alpha^i)$, are masked by (Alg.~\ref{algo:proc}-line \ref{line:6}). 
Overall, for a DNN, the exact quantities enumerated in \S\ref{section:threat}
are stored in the enclave $\mathcal{E}$. 
\subsection{Relation with BPDA}
\label{subsection:bpda}
\begin{figure}[!t]
    	\begin{center}
		\includegraphics[scale=0.6,trim={0 0 0 0}]{{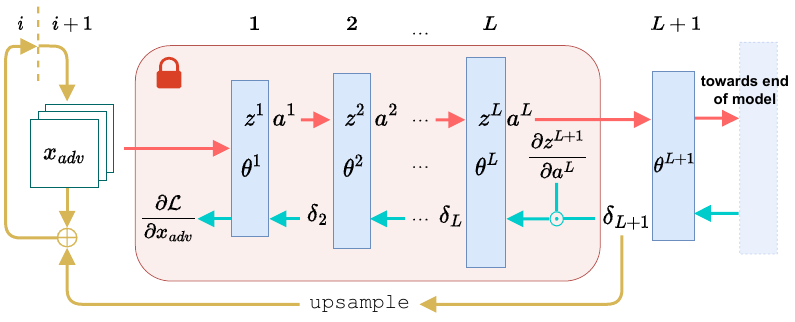}}
    	\end{center}
	\caption{Overview of the \sys defense scheme over the first layers of a machine learning model against an iterative gradient-based adversarial attack.
	Because the attacker cannot access operations in the first layers, he resorts to upsample his under-factored gradient $\delta_{L+1}$ to compute the gradient-based adversarial update.
	The figure depicts activations as transforms for a DNN.  \label{fig:shieldnn}}
\end{figure}

When training a DL model, a straight-through estimator allows a simple approximation of the gradient of a non-differentiable function (\eg a threshold)~\cite{DBLP:journals/corr/BengioLC13}.
In a setting discussed in~\cite{DBLP:conf/icml/AthalyeC018}, this idea is generalized to a so called Backward Pass Differentiable Approximation (BPDA) to illustrate how preventing a layer from being differentiable in hopes of defeating a gradient-based attack actually constitutes a fragile security measure.
In BPDA, the non-differentiable layer $f^l$ of a neural network $f^{1 \ldots L}(\cdot)$ would be approximated by a neural network $g$ \emph{s.t.} $g(x) \approx f^l(x)$ and be back-propagated through $g$ instead of the non-differentiable transform.
This method is what a malicious node adopts against \sys by upsampling the adjoint of the last clear layer $\delta_{L+1}$ to bypass the shielded layers. An illustrative case for a DNN is shown in Fig.\ref{fig:shieldnn}. However, the attacker operates with two limiting factors: \textit{(i)} in a real-world scenario, the attacker possesses limited time and number of passes before the broadcast model becomes obsolete; \textit{(ii)} we hypothesize the attacker does not have priors on the parameters of the first layers of the model, therefore the adversary is effectively left without options for computing the gradient-based update other than training a BPDA of the layer.
Although this attack makes a fundamental pitfall of gradient masking techniques, it is worth noting this step becomes increasingly difficult for the attacker as larger parts of the model are hidden from him since it would suppose he has training resources equivalent to that of the FL system. As a side note, we mention that there exist recent defenses against BPDA~\cite{9420266}.

In \S\ref{section:eval}, we investigate to what extent an evasion attack can be mitigated by \sys and whether the upsampling of the under-factored gradient (which is merely a linear transformation of the correct gradient in the case of the first transformation in many vision models) as a substitute for the masked backward process constitutes a possible last resort for the malicious node.
\section{Evaluation}
\label{section:eval}

\textit{What is the effect of forcibly removing leftmost factors from the chain-rule of an attacker conducting an adversarial attack?}
To answer this question, we conduct attacks on a robust ensemble model with shielded layers.
\subsection{ Evaluation Metric and Ensemble Defense Setup}
The \textit{clean accuracy} of a model refers to its standard test accuracy, to differentiate it from the context of an evasion attack where the goal of the defender is to score high on \textit{astuteness} (\ie \textit{robust accuracy}) against a test set with adversarial perturbation.
We chose as target model an ensemble of two state of the art high clean accuracy model trained on CIFAR-10~\cite{Krizhevsky09}: \emph{(i)} a Vision Transformer ViT-L/16~\cite{50650} and \emph{(ii)} a Big Transfer model BiT-M-R101x3~\cite{10.1007/978-3-030-58558-7_29} which stems from the CNN-based ResNet-v2 architecture~\cite{He2016Sep}. 
We chose an ensemble because, generally, when dealing with the image classification task, adversarial examples do not \textit{transfer} well between Transformer based and CNN based models~\cite{9710333}. This means that an example crafted to fool one type of model in particular will rarely defeat the other.
This allows for mitigating attacks that target either one of the two specifically, by exploiting a combination of the model outputs to maximize chances of correct prediction. 
In this paper, we use random selection~\cite{Srisakaokul2018Aug}, where, for each sample, one of the two models is selected at random to evaluate the input at test time. 

\textbf{\sys Shielding defense of the white-box.}
To simulate the shielding inside the TEE enclave, we deprive the attacker of the aforementioned quantities (\S\ref{ssec:peltashield}) in the two models separately, then jointly. 
For the ViT model, all transforms up to the position embedding~\cite{50650} are included. 
This means that the following steps occur inside the enclave: separation of the input into patches ${x}_p^n$, projection onto embedding space with embedding matrix $E$, concatenation with learnable token $x_{class}$ and summation with position embedding matrix $E_{pos}$:
$$
z_0=\left[x_{\text {class }} ; {x}_p^1 {E} ; {x}_p^2 {E} ; \cdots ; {x}_p^N {E}\right]+{E}_{\text {pos}}
$$

For the Big Transfer (BiT) model, the scheme includes the first weight-standardized convolution~\cite{10.1007/978-3-030-58558-7_29} and its following max-pooling operation.
Notice that, for both models, we obfuscate either two learnable transformations or a non-invertible parametric transformation in the case of MaxPool~\cite{Zeiler2014}, so the attacker cannot retrieve the obfuscated quantities without uncertainty.
Table~\ref{table:size} reports the estimated overheads of the shield for each setting.
The shielding of the ensemble requires up to 28.96~MB of TEE memory at worst, consistent with what typical TrustZone-enabled devices allow~\cite{amacher2019performance}.
\begin{center}
\begin{table}[!t]
	\centering
	\normalsize
	\setlength{\tabcolsep}{10pt}
	\begin{center}
		\rowcolors{1}{gray!0}{gray!10}
		\begin{tabularx}{\columnwidth}{lcc}
			\rowcolor{gray!50}
			\textbf{Model} & \textbf{Shielded portion} & \textbf{TEE mem. used}\\
			\rowcolor{gray!1}
			ViT-L/16  &   0.350\% & 12.01~MB \\
			BiT-M-R101x3   & 0.004\% & 16.95~MB \\
			\bottomrule
		\end{tabularx}
	\end{center}
		\vspace{3pt}
	\caption{\label{tab:size}Estimated enclave memory cost and model portion shielded in each setting.
	Values are shown for the worst case where intermediate activations and gradients inside the shield are not flushed after the back-propagation algorithm uses them.
	The ensemble value sums both models in the worst case where enclaves are not flushed between evaluation of either of the two models.}
	\label{table:size}
\vspace{-10pt}
\end{table}
\end{center}
\subsection{Attacker Setup} 

\label{subsection:saga}
To circumvent the ensemble defense, the attacker uses a non-targeted gradient-sign attack: the Self-Attention Gradient Attack~\cite{9710333}.
It iteratively crafts the adversarial example by following the sign of the gradient of the losses as follows:
\begin{equation}
\label{eq:fgsm}
    x_{a d v}^{(i+1)}=x_{a d v}^{(i)}+\epsilon_{s} * \operatorname{sign}\left(G_{b l e n d}\left(x_{a d v}^{(i)}\right)\right)
\end{equation}
where we initialize $x_{a d v}^{(1)}=x$ the initial image and $\epsilon_{{S}}=3.1\mathrm{e}{-3}$ is a given set attack step size (chosen experimentally).
Additionally, $G_{b l e n d}$ is defined as a weighted sum, each of the two terms working towards computing an adversarial example against either one of the two models:
\begin{equation}
    G_{b l e n d}\left(x_{a d v}^{(i)}\right)= \alpha_{k} \frac{\partial \mathcal{L}_{k}}{\partial x_{a d v}^{(i)}}+ \alpha_{v} \phi_{v} \odot \frac{\partial  \mathcal{L}_{v}}{\partial x_{a d v}^{(i)}}
\end{equation}
$\partial \mathcal{L}_k / \partial x_{a d v}^{(i)}$ is the partial derivative of the loss of the CNN-based architecture BiT-M-R101x3, and $\partial L_v / \partial x_{a d v}^{(i)}$ is the partial derivative of the loss of the Transformer-based architecture ViT-L/16.
Their prominence is controlled by the attacker through two manually set weighting factors, $\alpha_k=2.0\mathrm{e}{-4}$ and $\alpha_v=1-\alpha_k$.
For the ViT gradient, an additional factor is involved: the self-attention map term $\phi_v$, defined by a sum-product:
\begin{equation}
    \phi_{v}=\left(\prod_{l=1}^{n_{l}}\left[\sum_{i=1}^{n_{h}}\left(0.5 W_{l, i}^{(a t t)}+0.5 I\right)\right]\right) \odot x_{a d v}^{(i)}
\end{equation}
where $n_h=16$ is the number of attention heads per encoder block of the ViT model, $n_l=24$ the number of encoder blocks in the ViT model, $W_{l, i}^{(a t t)}$ is the attention weight matrix in each attention head and $I$ the identity matrix. $\odot$ is the element wise product.

\begin{table*}[!t]
	\centering
	\normalsize
	\setlength{\tabcolsep}{10pt}
	\begin{center}
		\rowcolors{1}{gray!0}{gray!10}
		\begin{tabularx}{1.9\columnwidth}{Xcccccc}
			\rowcolor{gray!50}
			\cellcolor{gray!0} & \multicolumn{2}{c|}{\textbf{Baseline}} & \multicolumn{4}{c}{\textbf{Applied Shield}}\\
			\rowcolor{gray!25}
			\textbf{Model Acc.} & \textbf{Clean} & \multicolumn{1}{c|}{\textbf{Random}} & \textbf{None} & \textbf{ViT-L/16} & \textbf{BiT-M-R101x3} & \textbf{Ensemble}\\
			\rowcolor{gray!1}
			ViT-L/16  &   99.5\% & \multicolumn{1}{c|}{99.5\%} & 28.1\% & 99.2\% & 14.1\% & 99.5\%\\
			BiT-M-R101x3   & 99.1\% & \multicolumn{1}{c|}{98.8\%} & 25.2\% & 0.3\% & 78.9\% & 98.5\%\\
			Ensemble  &   99.3\% & \multicolumn{1}{c|}{98.9\%} & 27.2\% & 49.7\% & 46.4\% & 98.8\%\\
			\specialrule{.8pt}{0pt}{2pt}
		\end{tabularx}
	\end{center}
		\vspace{3pt}
	\caption{\label{tab:results}Robust accuracy of a shielded ensemble against SAGA on CIFAR-10 (higher values favor the defender). Baseline values show clean accuracy, astuteness against random attack on the $l_{\infty}$ ball. Applied Shield values show per-model robust accuracy against different shielding setups.}
	\label{table:results}
\vspace{-10pt}
\end{table*}

Facing the \sys shielded setting, the attacker carries out the SAGA without the masked set $\left\{\partial f/\partial {x}\right\}^L$ (and the adjacent quantities otherwise enabling its unambiguous reconstruction).
It tries to exploit the adjoint $\delta_{L+1}$ of the last clear layer by applying to it a random-uniform initialized upsampling kernel.
This process, called \textit{transposed convolution}~\cite{Dumoulin2016AGT}, essentially is a geometrical transformation applied to the vector gradient at the backward pass of a convolutional layer.
Although this method does not offer any guarantee of convergence towards a successful adversarial example, it allows to understand whether the adjoint can still serve as a last resort when no priors on the shielded parts are available under limited resource constraint.

\subsection{Benchmarks and initial results}
\begin{figure}[!t]
	\begin{center}
		\includegraphics[scale=0.34,trim={0 0 0 0}]{{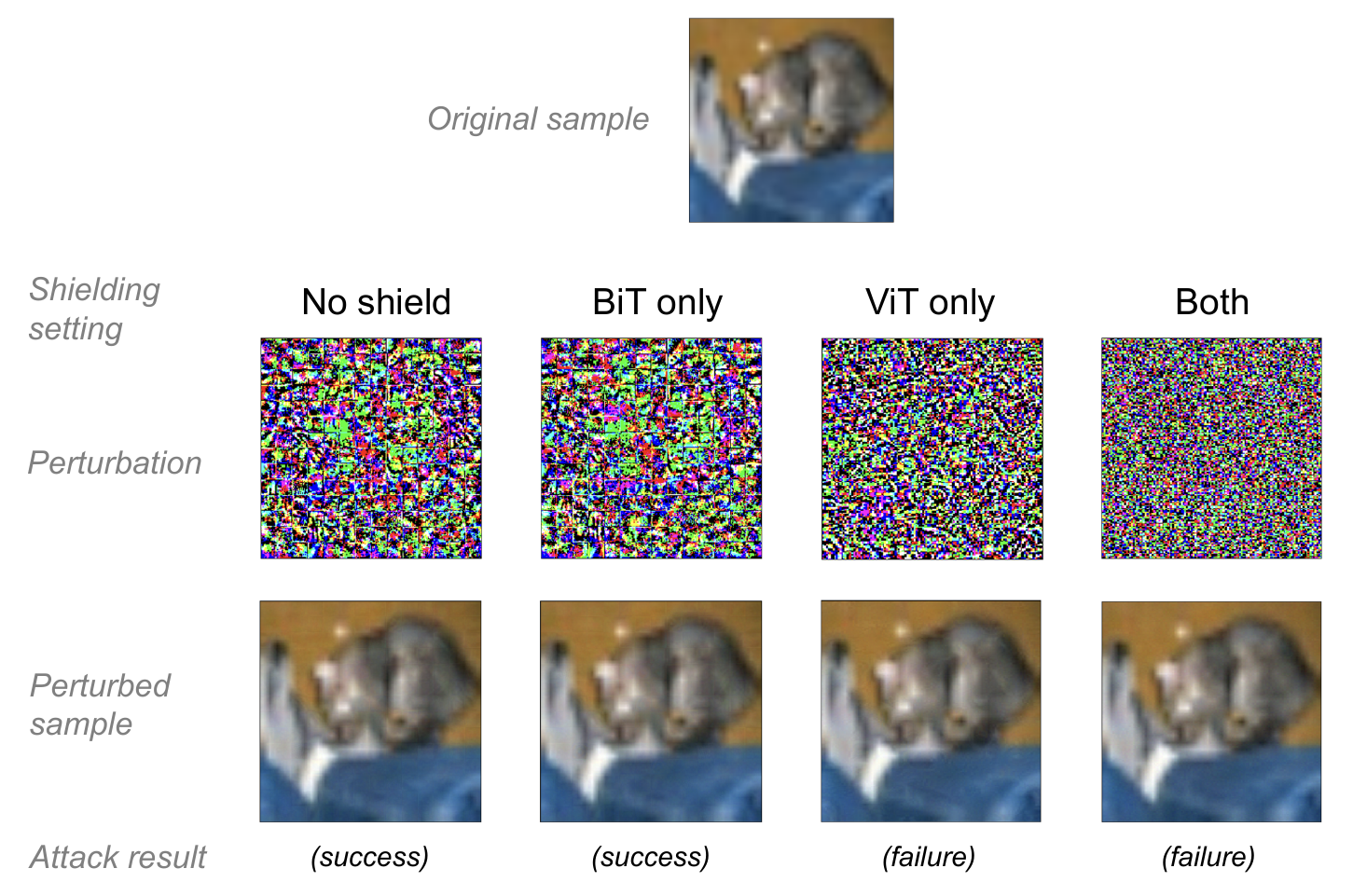}}
	\end{center}
	\caption{SAGA adversarial samples in four different shielding settings from an original sample. \label{fig:advex}}
\end{figure}

We select 1000 random samples from CIFAR-10~\cite{Krizhevsky09}, and evaluate the average robust accuracy of individual models and their ensemble over 10 attacks in each of the settings involving either BiT model with shield, ViT model with shield, both or none.
For each attack, the attacker makes ten passes ($i=1 $\mydots$10$ in Eq.~\ref{eq:fgsm}) of his adversarial instance through the ensemble to update it.
Table~\ref{table:results} shows our results.
For illustrative purposes, Fig.\ref{fig:advex} shows the generated perturbation on one sample in the different shielding settings.

\textit{Does \sys mitigate SAGA?}
We observe that the shielding greatly preserves the astuteness of individual models and of the ensemble, with $98.8\%$ robust accuracy ($1.2\%$ attack success rate).
However, individual model robust accuracies are not equally protected: ViT benefits more from \sys when applied only to it than BiT.
We explain this result as a general advantage in robustness of Transformer-based architectures over CNNs~\cite{DBLP:journals/corr/abs-2110-02797} with BiT being more sensitive to targeted attacks than its counterpart, and also more sensitive to adversarial examples crafted against the ViT. 
We further note that individual robust accuracies worsen compared to the non-shielded setting when only their counterpart is shielded. 
This is due to SAGA only moving the sample towards increasing the non-shielded loss at the expanse of its shielded counterpart, hence crafting adversarial samples that specifically target the non-shielded model.

\textit{Does upsampling constitute a last resort for the attacker?} Interestingly, the attack success rate of the upsampling against the \sys defense scheme surpasses that of the random attack against BiT when the shield is applied only to it.
We explain this behaviour as follows: contrary to the shielded layer in ViT that projects the input onto an embedding space, the last clear layer adjoint $\delta_{L+1}$ in the BiT still carries spatial information that could be recovered \eg through average upsampling.
These results suggest shielding both models for optimal security, given sufficient encrypted memory available in a target TEE.
\section{Conclusion and Future Work}
\label{section:conclu}
Adversarial attacks are difficult to defend against at inference time in FL, especially under white-box settings. 
We described how to mitigate such attacks by using hardware obfuscation to sidestep the white-box. We introduce a novel defense, \sys: a lightweight shielding method which relies on TrustZone enclaves to mask few carefully chosen parameters and gradients against iterative gradient-based attacks. 
Our evaluation shows promising results, sensibly mitigating the effectiveness of state of the art attacks. 

We intend to extend this work along the following directions.
Because the use of enclaves calls for somewhat costly normal-world to secure-world communication mechanisms, properly evaluating the speed of each collaborating device under our shielding scheme is needed on top of considering the aforementioned memory overheads (Table~\ref{table:size}) in order to assess the influence of \sys on the FL training's practical performance.
Additionally, a natural extension to this work is to apply \sys against other popular attacks, such as in Carlini \& Wagner~\cite{7958570}. 
\sys can also be extended to other gradient masking algorithms with backward approximation.
As mentioned in \S\ref{subsection:bpda}, it should also be explored that an attacker can \emph{(i)} exploit commonly used embedding matrices and subsequent parameters across existing models as a prior on the shielded layers (this case being circumvented by the defender if it trains its own first parameters) or \emph{(ii)} to train on their own premises  the aforementioned $g$ backward approximation (which needs not be of the same architecture as the shielded layers), although recent work shows limitation for such practice~\cite{sitawarin2022demystifying}.


\bibliographystyle{ACM-Reference-Format}

\bibliography{paper}

\end{document}